# 3D RECONSTRUCTION OF TEMPLES IN THE SPECIAL REGION OF YOGYAKARTA BY USING CLOSE-RANGE PHOTOGRAMMETRY

Adityo Priyandito Utomo[1], Canggih Puspo Wibowo[2]

[1] SMA Negeri 9 Yogyakarta
[2] Sagasitas Research Center
Jl. Sagan No.1, Terban, Gondokusuman, Yogyakarta, DIY, Indonesia, 55223[1]
Jl. Pogung Raya No 272F, Mlati, Sleman, DIY, Indonesia, 55284[2]
Email : adityo.priyandito@gmail.com[1], canggihpw@gmail.com[2]

**Abstract**
*Object reconstruction is one of the main problems in cultural heritage preservation. This problem is due to lack of data in documentation. Thus in this research we presented a method of 3D reconstruction using close-range photogrammetry. We collected 1319 photos from five temples in Yogyakarta. Using A-KAZE algorithm, keypoints of each image were obtained. Then we employed LIOP to create feature descriptor from it. After performing feature matching, L1RA was utilized to create sparse point clouds. In order to generate the geometry shape, MVS was used. Finally, FSSR and Large Scale Texturing were employed to deal with the surface and texture of the object. The quality of the reconstructed 3D model was measured by comparing the 3D images of the model with the original photos utilizing SSIM. The results showed that in terms of quality, our method was on par with other commercial method such as PhotoModeler and PhotoScan.*

## 1. Introduction

Special Region of Yogyakarta is a province in Indonesia, which has numerous temples [1,2]. Over the decades, structural damages have been occurring against the temples caused by earthquakes, volcanic ash, and other natural phenomena [1]. It has resulted in the degradation of the spiritual, aesthetic, and technical values that it holds, which is important from the perspective of culture and economy [3]. In order to restore those values, a reconstruction process needs to be done. However, lacking the representative documentation may result in an inaccurate reconstruction. Commonly, the government only takes 2D photographs of temples for the documentation, which lacks depth information on the temples. Thus, in this research, we present a method to create 3D objects of the temples by using close-range photogrammetry technique. There are two problems we would like to address in response to the status quo: the lack of depth information in the current 2D documentation (photographs) performed by the government and the expensive cost of the equipment available. This research has the objective of documenting temples in the Special Region of Yogyakarta in 3D representation, measuring the quality of the documentation result, and comparing the method presented to other existing methods of 3D documentation.

### 1.1. Close-Range Photogrammetry

Photogrammetry is the science of obtaining reliable information about the properties of surfaces and objects without physical contact with the objects, and of measuring and interpreting this information [4]. It can be considered to be an image processing technique where multiple photos taken from different angle were processed and triangulated to make a 3D representation of the object. Photogrammetry has been used to create the 3D representation of objects in some past works, such as the archeological site of Labna in Mexico [5], the statue of *Maddalena* in Italy [6], and Ivriz relief in Turkey [7]. However, all of them utilized high-cost commercial software such as PhotoModeler, Polyworks Modeler, and GeoMagic. In this research, we opted to use the low-cost alternative method.

Photogrammetry can be divided into two kinds based on the camera locations that are aerial and close-range photogrammetry. Aerial photogrammetry used to make a topographical map by taking photos from airplanes, helicopters, or drones. On the other hand, close-range photogrammetry focused on creating a 3D representation of smaller objects, such as buildings, structures, etc. Here, typically the camera is placed on a tripod or hand-held. In this research, we adapted close-range photogrammetry explained in [8].

### 1.2. Research Method

The process of photogrammetry applied here was an off-line system, meaning that the process was done in the laboratory after all images acquired. It has been reported that the off-line system has a better accuracy than the on-line one, which is done in real-time [8]. The method of this research was composed of three steps. First, images of temples were captured. Secondly, we applied various algorithms to process the images, such as keypoints extraction, feature description, point clouds construction, and densification. In the third step, surface and texture generation were employed as a post-processing step.

Finally, we performed some tests to find out the results of the photogrammetry. Each of those steps will be examined in the following subsections.

**1.2.1. Image Acquisition**

The image acquisition process used a non-metric consumer camera Nikon D3200, with the same focal length for all of the photos. The appropriate focal length was obtained through the process of calibration conforms to the unique characteristic of a non-metric camera. The scheme of the image acquisition process is shown in Figure 1. As showed in Figure 1, the image acquisition process must obey the parallax principle in which we had to move around the object to take multiple photos. Another thing that should be borne in mind is adjacent photos should have an overlapped area. This is necessary to ensure that 3D reconstruction can be done.

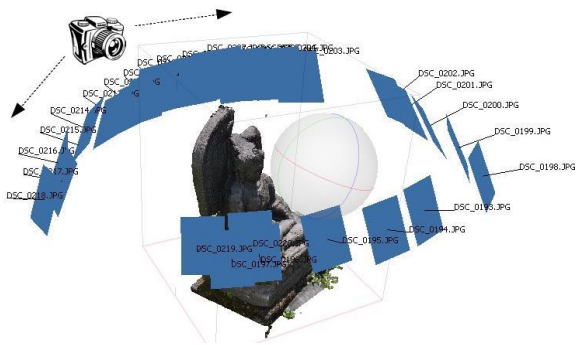

**Figure 1.** *Image Acquisition Process*

**1.2.2. Photogrammetry Process**

In order to get the correspondence between images, keypoints from each image have to be identified. Here, we used A-KAZE [9] to extract the keypoints. A-KAZE has been considered faster and having a better performance compared to state-of-the-art methods such as BRISK, ORB, SURF, and SIFT. It is an improved version of KAZE [10] with a reduced computational time. The main idea of A-KAZE and KAZE is the use of a nonlinear scale space, while the others commonly use the linear one such as Gaussian scale space. Then, keypoints were determined by computing the response of scale-normalized determinant of the Hessian matrix ($H_{norm}L$) at multiple scale levels described as

$$\det H_{norm}L = t^2 \left( L_{xx} L_{yy} - L_{xy}^2 \right) \quad (1)$$

where ($L_{xx}$, $L_{yy}$) are the second order horizontal and vertical derivatives respectively, $L_{xy}$ is the second order cross derivative, and $t^2$ indicates the scale parameter. Here, we selected the maxima in scale and spatial location of the Hessian to be the keypoints. Figure 2 shows the keypoints obtained after applying the A-KAZE algorithm.

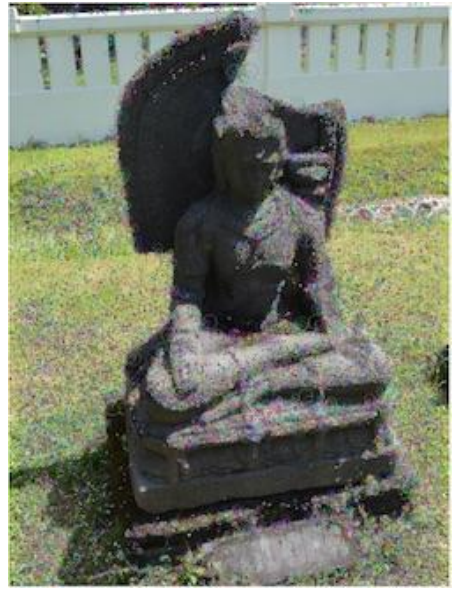

**Figure 2.** *Keypoint Detection of the "Arca Budha" Photo*

After the keypoints were obtained, feature descriptors were calculated. Local Intensity Order Pattern (LIOP) [11] was employed to extract information regarding the relationships among intensities of neighboring pixels. The procedure of LIOP is as follows:

1. Create local patches of an image with radius $R$ at keypoints.
2. Sort all pixels in each local patch in non-descending order. Then, the local patch is equally quantized into $B$ ordinal bins according to their orders.
3. For each pixel $x$ in the ordinal bin, LIOP is computed based on the intensity order of its neighboring pixels. The calculation is given in [11].
4. Construct the LIOP feature descriptor by concatenating the summation of LIOP in each ordinal bin as follows:

$$LIOP\ descriptor = (des_1, des_2, des_3, ..., des_B) \quad (2)$$

where,

$$des_i = \sum_{x \in bin_i} LIOP(x) \quad (3)$$

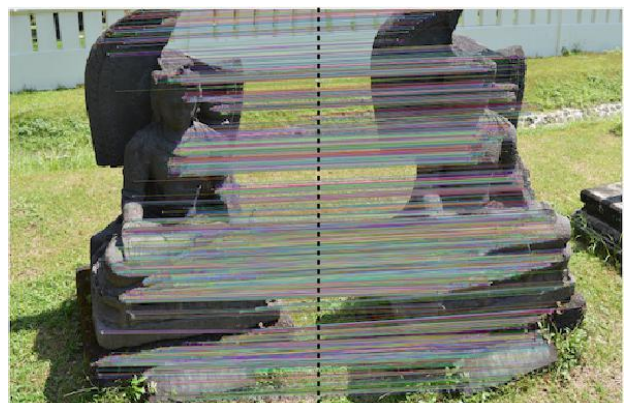

**Figure 3.** *Feature Matching of the "Arca Budha" Photos*

The descriptors that have been computed then used to obtain the correspondence between images through feature matching (Figure 3). Then, in order to construct point clouds, we estimated the camera orientations from the matching images. Here, rotation averaging approach was applied, meaning that all camera orientations are solved simultaneously from input pairwise relative rotations. The goal of rotation averaging is the creation of a set of rotation matrices as the representation of camera orientations. In this work, we employed L1 Rotation Averaging (L1RA) [12] which has been declared as significantly outperforms other methods. L1RA can be seen as an improvement of Lie-Algebraic Rotation Averaging [13] by adding a technique to deal with outliers. Rotation averaging produces sparse point clouds. In the next step, we used Multi-View Stereo (MVS) [14] to create a dense point cloud from the sparse one. The idea behind MVS is the use of a region growing approach. Here, successfully matched points were used to estimate its neighboring pixels. The example of sparse and dense point cloud is given in Figure 4. It can be observed that the geometry shape of the dense object is more detail due to the utilization of MVS.

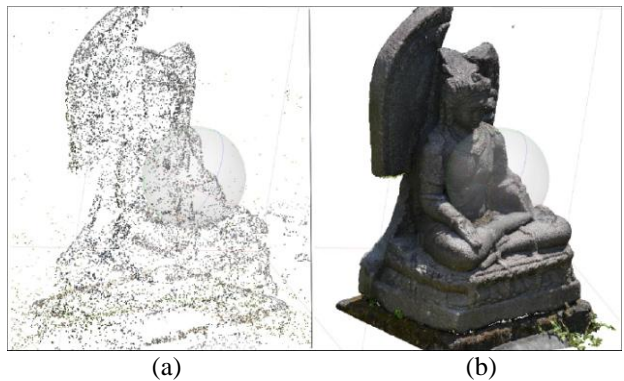

(a)  (b)

**Figure 4.** *Point Cloud of the "Arca Budha" Photo: (a) Sparse and (b) Dense.*

### 1.2.3. Post-processing

In order to cope with the surface of the temples, we used Floating Scale Surface Reconstruction (FSSR) [15] to generate geometric shapes from the dense point clouds. Afterward, Large Scale Texturing [16] was utilized to achieve texture maps from the 3D object, obtained through the image projections of the photos. In order to achieve a clean model fit for the current-generation physical based rendering, lighting information was removed from the final model texture using a high-pass filter. Additional maps such as specular, metalness, and bump was acquired by algorithmic means using several different techniques and algorithms. Finally, in order to bring the highest visual fidelity of the model we rendered them with Renderman 20 which is free for non-commercial use. Figure 5 shows the results of applying FSSR to generate the surface and Large Scale Texturing to create the texture of 3D object. It can be seen that surface and texture generation make the object more similar to the original.

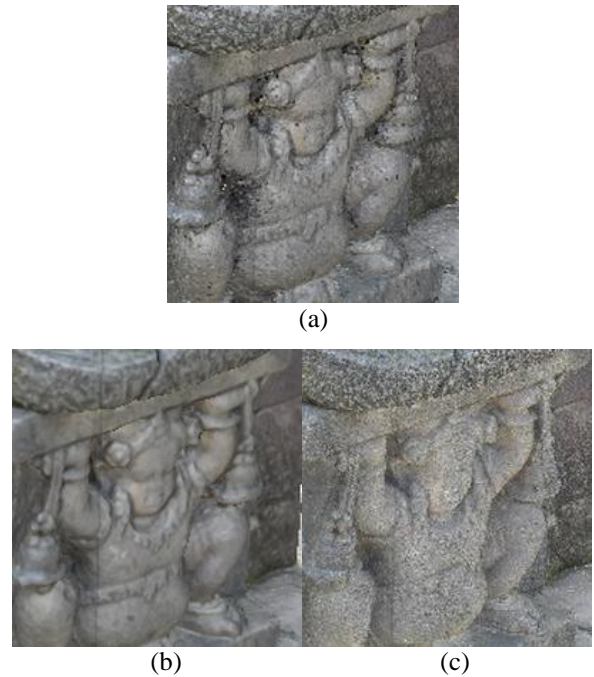

(a)

(b)  (c)

**Figure 5.** *Post-processing effect on (a) Dense Point Cloud: (b) After Surface Generation and (c) After Texture Generation.*

In this work, to implement the aforementioned photogrammetry algorithms, we used various free and open source libraries shown in Tabel 1. The use of free libraries can reduce the production cost. With the open source license, the libraries can be enhanced with community supports.

**Tabel 1.** *Free and Open Source Libraries Used to Implement the Algorithms*

| Algorithm | Library Implementation |
|---|---|
| A-KAZE | Akaze-eigen [17] |
| LIOP | VLFeat [18] |
| L1RA | OpenMVG [19] |
| MVS | MVE [20] |
| FSSR | MVE |
| Large Scale Texturing | MVS-Texturing [21] |

### 1.2.4. Quality Measurement

In order to validate the results, we measured image similarity between the original image and its 3D representation in the same angle. Here, SSIM (Structural Similarity Index) [22] was employed to measure the image similarity based on its luminance, contrast, and structure. Suppose **x** and **y** are two image signals with the same size. The SSIM of **x** and **y** is defined as

$$\text{SSIM}(\mathbf{x},\mathbf{y}) = \frac{(2\mu_x\mu_y + c_1)(2\sigma_{xy} + c_2)}{(\mu_x^2 + \mu_y^2 + c_1)(\sigma_x^2 + \sigma_y^2 + c_2)} \quad (4)$$

where $\mu_x$, $\mu_y$, $\sigma_x^2$, $\sigma_y^2$, and $\sigma_{xy}$ are the average of **x**, average of **y**, variance of **x**, variance of **y**, and covariance of **x** and **y**, respectively. While $c_1$ and $c_2$ are the variables employed to stabilize the division with a weak denominator. SSIM has scales goes from -1 up to 1 (where 1 indicates that the pair is completely identical images).

## 2. Discussion

We collected photos from five temples in the Special Region of Yogyakarta that are Lumbung, Barong, Prambanan, Sambisari, and Sewu temples with photo resolution 4000 x 6016. In total, 1319 photos were taken representing 60 objects in the temples. Details of number for each temple are shown in Table 2. All photos were able to be reconstructed in 3D representations. Thus, totally there are sixty 3D representations produced. To avoid ambiguity, photos resulted from 3D representation will be called "3D photo" and the photos acquired will be called "original photo" onwards.

**Table 2.** *Number of Original Photos Collected for each Temple*

| Temple | Number of Object | Total Original Photos |
|---|---|---|
| Sambisari | 12 | 315 |
| Barong | 11 | 310 |
| Prambanan | 22 | 403 |
| Lumbung | 3 | 60 |
| Sewu | 12 | 231 |
| Total | 60 | 1319 |

In order to ensure the quality of the 3D photos, we performed two kinds of test. The first test was carried out to know the quality of the 3D photos based on its similarity to the original photos. Then, in the second test, we compared our methods to other existing commercial methods. Both tests were done in a computer with processor Intel i3 4350, 3.5 GHz of clock speed, 4 GB of RAM, equipped with graphic card Nvidia® Geforce™ GTX 750 Ti, and Ubuntu OS 16.04.

### 2.1. Quality of the 3D Photos

In this test, 15 sets of 3D photos were selected. Each set is the product of an object in the temple. For each 3D photo, we measured the similarity to the original photo by computing the SSIM value. The results are presented in Table 3. It can be seen that in 0.79 value of average SSIM, our method produces a quite good 3D representation of temples. The value indicates that only a small part of the visual information was lost during the 3D reconstruction process. From Table 3, we can also infer that a number of original photos used do not contribute much towards the final quality of the 3D representation.

**Table 3.** *SSIM Averages for Different Number of Original Photos*

| Photos Set | Number of Original Photos | SSIM Average |
|---|---|---|
| Set 1 | 34 | 0.77 |
| Set 2 | 28 | 0.78 |
| Set 3 | 15 | 0.78 |
| Set 4 | 20 | 0.79 |
| Set 5 | 53 | 0.81 |
| Set 6 | 24 | 0.79 |
| Set 7 | 18 | 0.81 |
| Set 8 | 19 | 0.81 |
| Set 9 | 8 | 0.79 |
| Set 10 | 6 | 0.79 |
| Set 11 | 9 | 0.77 |
| Set 12 | 61 | 0.78 |
| Set 13 | 25 | 0.81 |
| Set 14 | 25 | 0.80 |
| Set 15 | 20 | 0.79 |
| Average from All Sets | | 0.79 |

### 2.2. Comparison with Other Methods

In order to see our method's standing among currently available high-cost methods (PhotoModeler and PhotoScan), we measured the performance of each method. Here, we take into consideration the average of SSIM value, time spent, CPU usage, and memory (RAM) usage. Due to the limitation of our computation equipment in this test we only used a set of original photos. We created two sets from it with varying the photo resolution and the number of original photos. The details of the photo sets are described in Table 4.

**Table 4.** *Photo Sets used in Comparison Test*

| Photos Set | Photos Resolution | Number of Original Photos |
|---|---|---|
| Set A | 4000 x 6016 | 16 |
| Set B | 500 x 752 | 30 |

The performance comparison is given in Table 5. In terms of SSIM value, there is no significant difference between our method and other high-cost methods. It means that our method is in a position to producing 3D representations on par with high-cost methods.

**Table 5.** *Performance Comparison with Other Methods*

| Methods | Average | | | |
| | SSIM value | Time spent (s) | CPU usage (%) | RAM usage (MB) |
|---|---|---|---|---|
| PhotoModeler | 0.74 | 3490 | 20.40 | 552 |
| PhotoScan | 0.73 | 3534 | 22.14 | 552 |
| Our Method | 0.72 | 4102 | 28.55 | 729 |

Furthermore, if we take a look at the time spent, CPU usage, and RAM usage, our method is falling behind the other two. However, the difference in these parameters can be considered not essential in our work since we are more focused on producing a good quality of 3D representations.

## 3. Conclusion

We presented a method to document temples in the Special Region of Yogyakarta in the form of 3D representation by using close-range photogrammetry. Despite the drawbacks in reconstruction time, CPU, and RAM usage, our method is still advisable to be used to construct 3D objects due to the good quality of results. Based on the SSIM values, our method yields a good quality which is on the same level with the high-cost methods available in the market.

**Acknowledgement**

This research was funded by Senior High School Section of Department of Education, Youth, and Sports of Yogyakarta Special Region, Indonesia under the Senior High School Science Clinic Program 2016.

**Biography of Authors**

*Adityo Priyandito Utomo,* currently studying at SMAN 9 Yogyakarta Indonesia, taking the natural science program and now at the eleven grade. His field of interest including but not limited to: machine learning, neural network, social complexities, computer imagery, and data visualization.

*Canggih Puspo Wibowo,* received a bachelor of engineering degree (S.T.) in electrical engineering from Universitas Gadjah Mada Indonesia in 2011. Then, in 2014 he received a master of engineering degree (M.Eng.) from King Mongkut's Institute of Technology Ladkrabang Thailand. Currently, he is a mentor of computer science research group at Sagasitas. His research interests are focused on pattern recognition and biometrics signal processing.